\documentclass[11pt]{article}

\usepackage{amsmath,amsfonts,bm}

\newcommand{\uiuc}[1]{#1$^1$}
\newcommand{\aws}[1]{#1$^2$}
\newcommand{\nvidia}[1]{#1$^3$}
\newcommand{\meta}[1]{#1$^4$}

\usepackage{xspace}
\usepackage{xcolor}
\usepackage{graphicx}
\usepackage{nicematrix}

\newcommand\Comment[1]{}

\newcommand\llm{LLM\xspace}
\newcommand\tir{TIR\xspace}
\newcommand\llmfull{Large Language Model\xspace}
\newcommand\tirfull{Tool-Integrated Reasoning\xspace}
\newcommand{\grpo}{GRPO\xspace} %
\newcommand{\grpofull}{Group Relative Policy Optimization\xspace} %
\newcommand{\ours}{GTPO\xspace} %
\newcommand{\oursfull}{Group Turn Policy Optimization\xspace}

\newcommand{\Rl}{RL\xspace}

\newcommand{\ie}{\emph{i.e.,}\xspace}

\def\eqref#1{equation~\ref{#1}}

\def\1{\bm{1}}

\DeclareMathAlphabet{\mathsfit}{\encodingdefault}{\sfdefault}{m}{sl}
\SetMathAlphabet{\mathsfit}{bold}{\encodingdefault}{\sfdefault}{bx}{n}

\usepackage[preprint]{acl}

\usepackage{times}
\usepackage{latexsym}

\usepackage[T1]{fontenc}

\usepackage[utf8]{inputenc}

\usepackage{microtype}

\usepackage{inconsolata}

\usepackage{graphicx}

\usepackage{hyperref}
\usepackage{url}
\usepackage{cases}
\usepackage{enumitem}
\usepackage{booktabs}
\usepackage{arydshln}
\usepackage{bbding}
\usepackage[dvipsnames, table]{xcolor}
\usepackage{pifont}
\usepackage{tabularx}
\usepackage{caption}
\usepackage{subcaption}
\usepackage{algorithm}
\usepackage{algpseudocode}

\newcommand{\xmark}{\ding{55}}%
\newcommand{\kr}[1]{}
\newcommand{\sh}[1]{}
\newcommand{\hl}[1]{}

\title{Empowering Multi-Turn Tool-Integrated Agentic Reasoning with \\ \oursfull}

\author{
\noindent\begin{tabular}{>{\centering\arraybackslash}p{.152\textwidth}>{\centering\arraybackslash}p{.166\textwidth}>{\centering\arraybackslash}p{.182\textwidth}>{\centering\arraybackslash}p{.171\textwidth}>{\centering\arraybackslash}p{.17\textwidth}}
\uiuc{Yifeng Ding}\thanks{Work done at AWS AI Labs.} & \aws{Hung Le} & \nvidia{Songyang Han}\footnotemark[1] & \aws{Kangrui Ruan} & \nvidia{Zhenghui Jin}\footnotemark[1] \\  & \aws{Varun Kumar} & \meta{Zijian Wang}\footnotemark[1] & \aws{Anoop Deoras} & 
\end{tabular}\\
$^1$University of Illinois Urbana-Champaign \quad $^2$AWS AI Labs \quad $^3$NVIDIA \quad $^4$Meta \\
\texttt{yifeng6@illinois.edu} \quad \texttt{\{songyangh,bjin\}@nvidia.com} \\  \texttt{zijianwang@meta.com} \quad \texttt{\{kangruir,kuvrun,adeoras\}@amazon.com} 
}

\begin{document}
\maketitle
\begin{abstract}
Training \llmfull{s} (\llm{s}) for multi-turn \tirfull (\tir) -- where models iteratively reason, generate code, and verify through execution -- remains challenging for existing reinforcement learning (\Rl) approaches. Current \Rl methods, exemplified by \grpofull (\grpo), suffer from coarse-grained, trajectory-level rewards that provide insufficient learning signals for complex multi-turn interactions, leading to training stagnation.
To address this issue, we propose \textbf{\oursfull} (\textbf{\ours}), a novel \Rl algorithm specifically designed for training \llm{s} on multi-turn \tir tasks.
GTPO introduces three key innovations: (1) \textit{turn-level reward assignment} that provides fine-grained feedback for individual turns, (2) \textit{return-based advantage estimation} where normalized discounted returns are calculated as advantages, and (3) \textit{self-supervised reward shaping} that exploits self-supervision signals from generated code to densify sparse binary outcome-based rewards.
Our comprehensive evaluation demonstrates that \ours outperforms \grpo by 3.0\% across diverse math reasoning benchmarks, establishing its effectiveness. \ours also improves \grpo by 3.9\% on commonsense reasoning and program synthesis tasks, demonstrating its generalizability to non-math domains. Importantly, \ours incurs negligible overhead, ensuring its practicality for real-world scenarios.

\end{abstract}

\section{Introduction}
\begin{figure*}[htbp]
\centering
\includegraphics[width=0.95\linewidth]{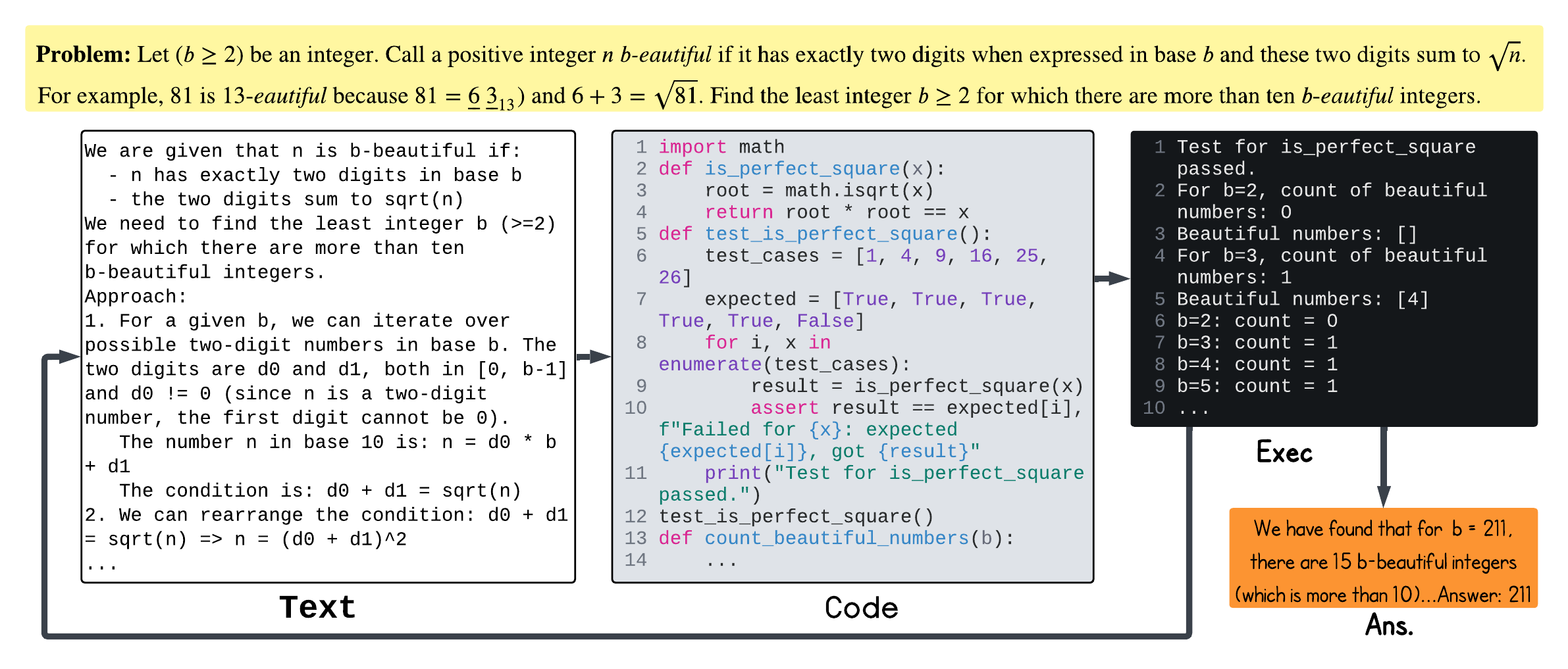}
\caption{
\textbf{Tool-integrated reasoning (TIR)}: Given a problem, the model progresses over multiple turns, where each turn consists of: (1) generating textual reasoning, (2) invoking tools (e.g., code), and (3) incorporating tool execution results to refine its understanding. 
The model repeats this cycle until a termination condition is met, either by producing a final answer or by reaching a predefined stopping criterion.
}
\label{fig:example}
\vspace*{-\baselineskip}
\end{figure*}

Reinforcement learning (RL) has become a powerful training technique to improve language model reasoning capabilities, enabling these models to generate long and complex chains of thoughts \citep{jaech2024openai, qwq32b, deepseekr1}.
To improve model reasoning beyond its natural language form, recently \citet{jin2025search, feng2025retoolreinforcementlearningstrategic} adopted tool-using strategies \citep{chen2023program, pmlr-v202-gao23f} and optimized language models for tool-integrated reasoning (TIR). 
In domains that require intense and symbolic reasoning, TIR can facilitate precise and numerical validations between intermediate reasoning steps \citep{hendrycks2021measuring, he-etal-2024-olympiadbench}.
Figure \ref{fig:example} provides a motivating example task with TIR.

In TIR, the integration of tools provides an executable interface a model can interact with across multiple turns.
In each turn, the model can iteratively evoke tools, receive tool output results, and revise its reasoning accordingly. 
As this multi-turn extension inherently complicates LLM reasoning trajectories, we observed severe issues when applying advanced RL algorithms such as Group Relative Policy Optimization (GRPO) and its variants \citep{deepseek-math, liu2025understanding, yu2025dapo} for TIR.
Specifically, we observed empirically that models' performance often stops improving effectively when training with GRPO for multi-turn \tir, even with continued learning iterations.

We identify two major challenges in existing RL approaches for TIR tasks. 
Firstly, current RL algorithms such as GRPO \citep{deepseek-math} adopt a simple \textit{sequence-level} 
reward assignment, and represent the advantage of each token using the normalized sequence-level \textit{reward}. 
While scalable, this assignment strategy introduces arbitrary and noisy reward feedback in multi-turn TIR. 
Specifically, depending on the tool execution outputs in each turn, the model can reflect and revise its reasoning steps significantly in the subsequent turns. 
Potentially, this dynamic model behavior can drastically shift the underlying reward contribution across all reasoning turns.
In this case, we found that a \textit{turn-wise} reward assignment strategy along with \textit{return}-based advantage is more appropriate. 

Furthermore, we observe that existing work often leverages a simple binary outcome reward 
based on the accuracy of the final response \citep{deepseekr1, feng2025retoolreinforcementlearningstrategic}. 
While accurate and efficient, such a binary sparse reward quickly becomes insufficient to bring enough learning signals in RL training for multi-turn \tir tasks.
Specifically, simply assigning zero to incorrect trajectories neglects the fact that trajectories with wrong final answers may still be partially correct and contain valuable learning signals, thus making RL training signals too sparse for \llm{s} to learn multi-turn \tir well.
Clearly, it's important to explore new reward shaping techniques that can densify the sparse binary outcome reward while maintaining its accuracy and efficiency.

To address the above challenges, we propose \textbf{\oursfull} (\textbf{\ours}).
Unlike existing approaches that rely on trajectory-level rewards \citep{feng2025retoolreinforcementlearningstrategic}, we introduce a fine-grained turn-level reward function that assigns diverse distinct rewards for individual turns within each trajectory.
Building upon research in conventional multi-turn RL \citep{shani2024multiturn, gao2025regressing}, we enable turn-level discounting to calculate return-based advantages.
Essentially, this reward assignment strategy and return-based advantage addressed not only the non-uniformity of reward distributions in TIR but also the temporal shift of rewards throughout the reasoning process. 

Additionally, \ours introduces a novel reward shaping technique that leverages self-supervision signals based on tool-calling contents in each turn. 
Specifically, different from directly assigning zero reward to negative trajectories in outcome-based reward, we leverage the accumulated code contents in \tir trajectories to compute mean similarity scores between negative and positive trajectories during training, and use such similarity scores as the partial rewards for negative trajectories.
We found this reward modeling strategy simple yet effective in augmenting conventional sparse binary outcome-based rewards and maintaining their original accuracy, without losing any noticeable efficiency.

Our comprehensive evaluation demonstrates that \ours achieves 3.0\% relative improvement
over \grpo on five diverse math reasoning benchmarks
. \ours also improves \grpo by 3.9\% on commonsense reasoning and program synthesis tasks, demonstrating its generalizability to non-math domains. Furthermore, \ours incurs negligible overhead, establishing \ours as a promising solution for the real world.

\section{Related Work}

\begin{table*}[t]
\centering
\small
\begin{tabular}{lccc}
\hline
\textbf{Feature / Dimension} & \textbf{GTPO (Ours)} & \textbf{GRPO / DAPO / GSPO} & \textbf{ReTool / ToRL / Search-R1} \\
\hline
MDP Formulation 
& {\color{Green} \CheckmarkBold}\ Turn-level
& {\color{Red} \xmark}\ Trajectory-level 
& {\color{Red} \xmark}\ Trajectory-level \\

Reward Granularity 
& {\color{Green} \CheckmarkBold}\ Dense (Self-Supervised Shaping)
& {\color{Red} \xmark}\ Sparse (Binary)
& {\color{Red} \xmark}\ Sparse (Binary) \\

Advantage Estimate
& {\color{Green} \CheckmarkBold}\ Discounted Return 
& {\color{Red} \xmark}\ Broadcast Uniform Reward
& {\color{Red} \xmark}\ Broadcast Uniform Reward \\
\hline
\end{tabular}
\caption{\textbf{A comparison between \ours and prior works:} different from prior works, \ours improves multi-turn RL training by introducing turn-level MDP formulation (\S\ref{sec:31}), return-based advantage estimates (\S\ref{sec:32}), and denser non-binary rewards based on self-supervised reward shaping (\S\ref{sec:33}).}
\label{tab:related_work}
\vspace*{-\baselineskip}
\end{table*}

Related to our work is the research of reasoning abilities in \llmfull{s} (\llm{s}). 
Early research work have demonstrated critical model behaviors such as reflection, deliberation, and correction \citep{wei2022chain, shinn2023reflexion}. 
\citet{yao2023tree, snell2024scaling} extended this line of research with inference-time scaling, in which sophisticated search techniques or nontrivial computational resources are deployed to control and manipulate model behaviors during reasoning processes. 
More related to our work is the research for model training techniques to improve model reasoning \citep{zelikman2022star, wang2024mathcoder, trung-etal-2024-reft, xie2024monte, kumar2025training}. 
Recently, \citet{qwq32b, deepseekr1, team2025kimi} showed that with appropriate reward functions and scalable RL training, \llm{s} can learn from its past reasoning, improving the reasoning qualities with naturally emerging behaviors and solving very complex tasks. 

Also related to our work is the research on tool-integrated reasoning (TIR). 
\citet{chen2023program, pmlr-v202-gao23f} proposed test-time strategies to integrate tools to solve mathematical problems. 
These tools can provide precise numerical validation at each step. 
More related to our work is the research on RL training strategies for TIR.
Retool~\cite{feng2025retoolreinforcementlearningstrategic}, ToRL~\cite{li2025torlscalingtoolintegratedrl}, and Search-R1~\cite{jin2025search} used RL with a binary and trajectory-level reward function based on the accuracy of final answers. 
Extending from these approaches, we introduce \ours with carefully designed turn-level reward assignment, discounted return-based advantage, and self-supervised reward shaping strategies to address the multi-turn nature and sparse reward feedback in \tir tasks. 
We conduct a systematic comparison of \ours and related works in Table \ref{tab:related_work} and Appendix \ref{sec:appendix_comparison}.

\section{Background}

\paragraph{Preliminaries.} 
We define a language model parameterized by $\theta$ as a policy $\pi_\theta$.
We denote $x$ as input to the model.
The likelihood under $\pi_\theta$ to obtain a response $y$ from $x$ is:
$\pi_\theta (y | x)=\prod_{t=1}^{|y|} \pi_\theta (y_t | x, y_{<t} )$
where $|y|$ is the number of tokens in $y$, $y_t$ denotes the $t$-th token in $y$, and $y_{<t}$ represents the part of $y$ before the $t$-th token.
Typically, in reasoning tasks, a verifier $v$ is available to assess the accuracy of the generated answer, resulting in a reward $r$. 
A simple verifier provides binary rewards $ r \in \{0, 1 \}$ where $1$ is for an answer that exactly matches the ground truth, and $0$ otherwise.

\paragraph{Tool-integrated reasoning (TIR).}
TIR \citep{chen2023program, pmlr-v202-gao23f} enhances language models with external tools to improve reasoning capabilities. 
TIR naturally decomposes output $y$ as a sequence of $n$ ``turns'': $y=\{y_1,b_1,y_2,b_2,\cdots,y_n\}$ and each turn $y_j \ (j<n)$ can be represented by $y_j=\{t_j,c_j\}$, where $t_j$ is the natural language reasoning, $c_j$ as the tool invocation, and $b_j$ as the feedback of tools execution.
The last turn $y_n$ will consist of natural language only: $y_n=t_n$, in which the final answer is extracted.

\paragraph{Group Relative Policy Optimization (\grpo).}
Recently, \citet{deepseek-math} proposed Group Relative Policy Optimization (\grpo) to scale RL training by removing the need to train and maintain a value model to estimate the rewards. 
Specifically, \grpo adopts group-based advantage estimation with the following objective \footnote{For notation, we simplify the details of data distributions under $\mathbb{E}$: ${ x \sim \mathcal{D},\, \{y_i\} \sim \pi_{\theta_\text{old}}}$ where $\mathcal{D}$ is the training dataset. Different from vanilla GRPO, we include several improvements DAPO~\cite{yu2025dapo} introduces in the objective.}: 
\vspace*{-1.1\baselineskip}
\begin{equation}
\begin{aligned}
& \mathcal{J}_\text{\grpo}(\theta) =  \mathbb{E}_{ x, \{y_i\}} 
[ \frac{1}{\sum_{i=1}^{G}|y_i|} \sum_{i=1}^{G} \sum_{t=1}^{|y_i|} \min \\
&\left(w_{i,t}(\theta) A_{i,t}, \, \mathrm{clip} \left( w_{i,t}(\theta), 1 - {\varepsilon_{\text{low}}}, 1 + {\varepsilon_{\text{high}}}\right) A_{i,t} \right) ] \label{eq:grpo}
\end{aligned}
\end{equation}
where $G$ is the number of responses generated from \llm for an input query $x$, $\varepsilon$ is a clipping range, and $w_{i,t}(\theta)$ is the importance ratio of the token $y_{i,t}$:
\vspace*{-\baselineskip}
\begin{align}
w_{i,t}(\theta) = \frac{ \pi_{\theta} (y_{i,t} | x, y_{i,<t}) }{ \pi_{\theta_\text{old}} (y_{i,t} | x,y_{i,<t})} \label{eq:importance_ratio}
\end{align}
Unlike PPO \citep{ouyang2022training}, $A_{i,t}$ in GRPO is measured as the group-based relative advantage:
\begin{align}
A_{i,t} = A_{i} = \frac{r(x, y_i) - \mathrm{mean} \left( \{ r(x, y_i) \}_{i=1}^G \right) }{ \mathrm{std} \left( \{ r(x, y_i) \}_{i=1}^G \right) }, \label{eq:grpo_advantage}
\end{align}
where each token $y_{i,t}$ within the same output $i$ shares the same advantage $A_{i}$. Following existing work~\citep{yu2025dapo}, we remove KL regularization from the training objective in our experiments.

\section{\oursfull}
\begin{figure*}[t]
\centering
\includegraphics[width=0.95\linewidth]{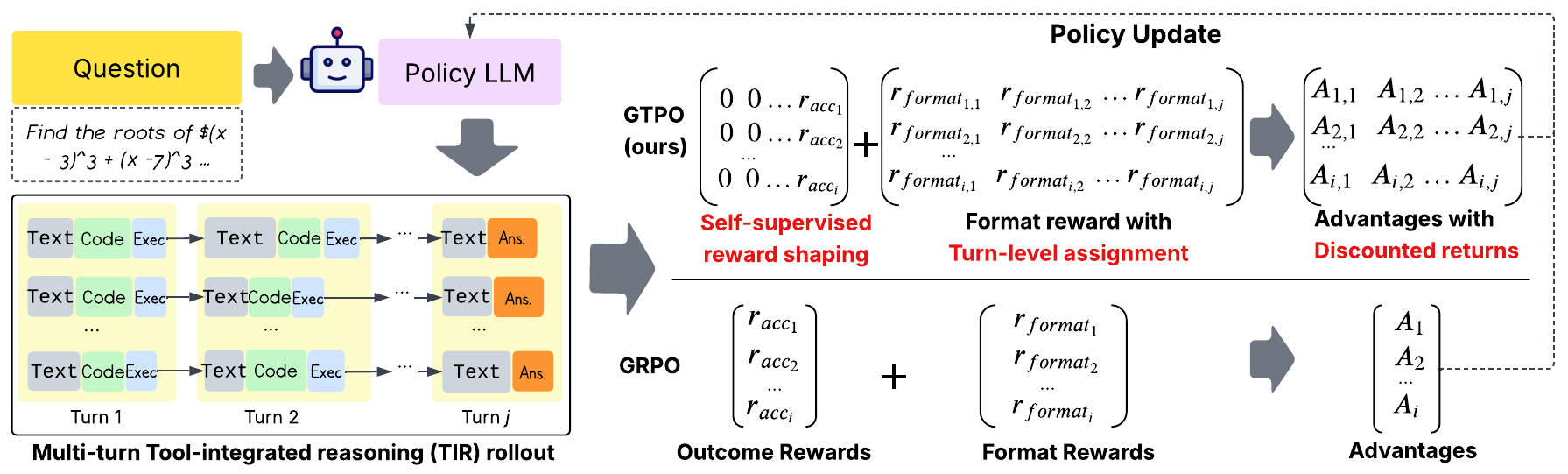}
\caption{
\textbf{An overview of \ours}: 
Unlike existing approaches that rely on trajectory-level rewards, \ours introduces a turn-level reward function that assigns diverse, rule-based rewards for individual turns within each trajectory and performs turn-level return-based discounting for advantage calculation. 
}
\vspace*{-0.8\baselineskip}
\label{fig:overview}
\end{figure*}

\begin{figure*}[!ht]
\centering
\includegraphics[width=0.95\linewidth]{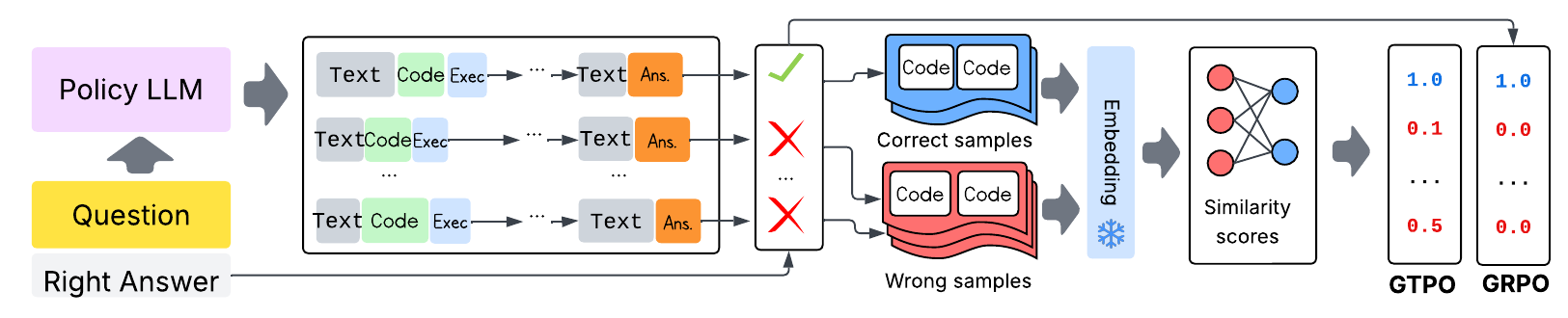}
\caption{
\textbf{GTPO reward shaping strategy}:
In \ours, each rollout trajectory is partitioned by final outcome (correct vs. incorrect), and the code content is extracted.
For each trajectory in the incorrect group, we compute its average similarity against all samples in the correct group and use the similarity score as its partial reward, so that wrong trajectories can still be properly utilized during training for more learning signals.
}
\label{fig:reward_shaping}
\vspace*{-\baselineskip}
\end{figure*}

Motivated by \grpo’s oversight of the multi-turn nature of \tir tasks, we propose \oursfull (\ours) to better fine-tune \llm{s} through the following strategies: 
(i) \emph{turn-level reward assignment} (\S\ref{sec:31}) assigns individual reward $r_i$ to individual turn $y_i$;
(ii) \emph{advantage with discounted return} (\S\ref{sec:32}) is measured by a discount factor $\gamma$ based on the corresponding turn-wise position; and
(iii) \emph{self-supervised reward shaping} (\S\ref{sec:33}) augments conventional binary reward models with a self-supervised scoring method.
Figure~\ref{fig:overview} illustrates the overall framework, and Figure \ref{fig:reward_shaping} details the reward shaping mechanism. In all the figures and equations, we use \textcolor{red}{red} to indicate the specific terms that differ between GRPO and \ours. We provide pseudocode (Algorithm \ref{alg:gtpo} in the Appendix) to further clarify the workflow of \ours.

\subsection{Turn-level Reward Assignment}\label{sec:31}
In the common practice of \grpo~\citep{deepseekr1, feng2025retoolreinforcementlearningstrategic}, a single terminal reward $r_i$ is assigned to a whole \tir trajectory $i$:
\vspace*{-0.6\baselineskip}
\begin{align}
r_i = \min(r_{\text{acc}_i}, r_{\text{format}_i}) \label{eq:grpo_reward}
\end{align}
\vspace*{-\baselineskip}
\begin{equation*}
\begin{split}
r_{\text{acc}_i} & =  \left\{
\begin{array}{cl}
  0   & {\text{if final answer is wrong}} \\
  1   & {\text{otherwise}} 
\end{array}
\right., \\
r_{\text{format}_i} & =  \left\{
\begin{array}{cl}
  0   & {\text{if trajectory has a format error}} \\
  1   & {\text{otherwise}} 
\end{array}
\right.
\end{split}
\end{equation*}

\grpo defines \tir with the following MDP:
\vspace*{-0.2\baselineskip}
\begin{itemize}[leftmargin=1em]
    \setlength{\parskip}{1pt}
    \item \textbf{State:} the initial input prompt $x$ as the state $s$
    \item \textbf{Action:} the whole \tir trajectory $y_i=\{y_{i,1},b_{i,1},y_{i,2},b_{i,2},\cdots,y_{i,T_i}\}$ as one action $a_i$ ($T_i$ refers to the number of turns in $y_i$)
\end{itemize}
\vspace*{-0.2\baselineskip}
While straightforward, this sequence-level reward assignment may introduce arbitrary and noisy feedback in the long multi-turn trajectory in \tir. Based on this observation, we propose a turn-level reward assignment that measures individual reward $r_j$ at turn $j$. Correspondingly, instead of treating the \tir task as a simple bandit problem, we construct a new MDP as follows, where we treat each turn $y_j$ in the trajectory as a separate action:
\begin{itemize}[leftmargin=1em]
    \setlength{\parskip}{2pt}
    \item \textbf{State:} the input prompt of the $j$-th turn $x_j=\{x,y_{i,1},b_{i,1},\cdots,y_{i,j-1},b_{i,j-1}\}$ as the state $s_{i,j}$ in the $i$-th trajectory
    \item \textbf{Action:} the $j$-th turn of the $i$-th trajectory $y_{i,j}=\{t_{i,j},c_{i,j}\}$ as the action $a_{i,j}$
\end{itemize}

Specifically, extending from the reward Eq.\ref{eq:grpo_reward} from \grpo, we can define $r_{i,j}$ for each turn $j$ in one \tir trajectory $i$ as follows, and more details have been included in Appendix \ref{sec:appendix_reward}:
\vspace*{-0.8\baselineskip}
\begin{align}
r_{i,\textcolor{red}{j}} = r_{\text{acc}_{i,\textcolor{red}{j}}} + r_{\text{format}_{i,\textcolor{red}{j}}} \label{eq:gtpo_reward}
\end{align}
\vspace*{-\baselineskip}
\begin{equation*}
\begin{split}
r_{\text{acc}_{i,\textcolor{red}{j}}} & =  \left\{
\begin{array}{cl}
  0   & {\text{if $y_{i,j}$ is not the last turn}} \\
  0   & {\text{if $y_{i,j}$ is the last turn and final}} \\
    &   {\text{answer is wrong}} \\
  1   & {\text{otherwise}} 
\end{array}
\right., \\
r_{\text{format}_{i,\textcolor{red}{j}}} & =  \left\{
\begin{array}{cl}
  -0.1   & {\text{if $y_{i,j}$ contains format errors}} \\
  0   & {\text{otherwise}} 
\end{array}
\right.
\end{split}
\end{equation*}

\subsection{Advantage with Discounted Return}\label{sec:32}
While the turn-level reward assignment strategy accounts for more fine-grained rewards in individual turns, it does not account for the sequential order of turns. 
Motivated by traditional RL practice \citep{shani2024multiturn, gao2025regressing}, we propose to incorporate the temporal effect between turns through a discounting factor $\gamma$ in the reward formula:
\vspace*{-0.6\baselineskip}
\begin{align}
R_{i,j} = \sum_{m=j}^{T_i} \gamma^{m-j} r_{i,m},
\label{eq:gtpo_discount_r}
\end{align}

where $r_{i,m}$ refers to the reward of turn $y_m$ in trajectory $i$. Essentially, the discounting factor $\gamma$ can systematically discount the value of future rewards with a diminishing values turn-by-turn. 
At each turn $j$, a return (\ie reward-to-go) $R_j$ is the sum of individual discounted rewards from the current turn until the terminal turn.
From Eq.\ref{eq:grpo_advantage}, the \ours advantages are then updated as: 
\begin{align}
A_{i,j,t} = \textcolor{red}{A_{i,j}} = \frac{\textcolor{red}{R_{i,j}} - \mathrm{mean} \left( \{ \textcolor{red}{R_{i,j}} \}_{\forall i,j} \right) }{ \mathrm{std} \left( \{ \textcolor{red}{R_{i,j}} \}_{\forall i,j} \right) },
\label{eq:advantage_gtpo}
\end{align}
where the advantage normalization is performed globally: we aggregate the advantages from all turns across all sampled trajectories and compute a single mean and standard deviation over this pooled set. This global normalization scheme naturally handles trajectories of varying lengths, as each turn-level advantage is treated as an independent sample in the shared normalization pool.

This turn-wise reward assignment with the discounting strategy addresses not only the non-uniformity of reward distributions in multi-turn TIR but also the temporal shift of rewards across reasoning steps throughout the reasoning process.
We further study the relationship between the values of $\gamma$ and the model's performance after RL training. As shown in \S\ref{sec:eval_result}, choosing the right $\gamma$ is crucial to the final success of \ours training. 

\subsection{Self-supervised Reward Shaping}\label{sec:33}
\grpo estimates advantages from the binary outcome reward \citep{deepseekr1}. 
While straightforward, such a sparse reward cannot bring enough learning signals for effective RL training.
Specifically, simply assigning $r_\text{acc} \in \{0,1\}$ neglects the fact that trajectories with wrong final answers may still be partially correct and contain good supervision data (e.g., partially correct code). 
Assigning strict $r_\text{acc}=0$ to failed trajectories makes RL training signals too sparse for \llm{s} to learn well. As such, we propose a simple yet effective strategy to assign partial rewards for $r_{\text{acc}}$ of the failed trajectories. 
See Figure \ref{fig:reward_shaping} for an overview.

Specifically, given a rollout sample $i$, we first extract and concatenate the tool invocation (code) contents: $c_{i,0} \oplus c_{i,1} \oplus \cdots \oplus c_{i,T_i-1}$. 
From a group of $G$ samples, we filter for samples with the final predicted answer 
matching the ground truth answer and denote this set of positive samples as $\mathcal{P}$. 
We then compute the partial reward as the mean similarity score between a negative code sample against all positive code samples. 
Essentially, extending $r_\text{acc}{}_{i,j}$ when $y_{i,j}$ is the last turn from Eq.\ref{eq:gtpo_reward}:

\vspace*{-1.2\baselineskip}
\begin{multline}
{r_\text{acc}{}_{i,j}} = \frac{\alpha}{|\mathcal{P}|}\sum\limits_{p\in\mathcal{P}}\mathrm{sim}\!\Big(\!\bigoplus_{m<j} c_{i,m}, \bigoplus_{m<j} c_{p,m}\!\Big) \text{ if } i \not\in \mathcal{P}
\end{multline}

\vspace*{-1.2\baselineskip}

Note that $r_\text{acc}{}_{i,j}=1.0$ if $i \in \mathcal{P}$, and $\alpha$ is the upper bound hyperparameter of the partial rewards.
$\mathrm{sim}(.,.)$ returns a similarity score between two input components, which can be any efficient off-the-shelf embedding model. 
When $|\mathcal{P}|=0$ (\ie the rollout group has no correct samples), no partial rewards will be assigned.
Apart from the final result of the ground truth to obtain the positive set $\mathcal{P}$, our reward shaping strategy does not need additional external supervision to measure the turn-wise reward, demonstrating its simplicity and efficiency to apply.
In practice, we set $\alpha=0.5$, which achieves good results across benchmarks empirically.

\begin{table*}[htbp]
\centering
\small
\begin{tabular}{@{}lcccccc@{}}
\toprule
\textbf{Models}              & \multicolumn{1}{c}{\begin{tabular}[c]{@{}c@{}}\textbf{AIME 2024}\\ (\emph{avg@16})\end{tabular}} & \multicolumn{1}{c}{\begin{tabular}[c]{@{}c@{}}\textbf{AIME 2025}\\ (\emph{avg@16})\end{tabular}} & \multicolumn{1}{c}{\begin{tabular}[c]{@{}c@{}}\textbf{MATH 500}\\ (\emph{pass@1})\end{tabular}} & \multicolumn{1}{c}{\begin{tabular}[c]{@{}c@{}}\textbf{AMC 2023}\\ (\emph{avg@16})\end{tabular}} & \multicolumn{1}{c}{\begin{tabular}[c]{@{}c@{}}\textbf{SVAMP}\\ (\emph{pass@1})\end{tabular}} & \textbf{Average} \\ \midrule
\textbf{Qwen2.5-7B-Instruct} &         6.70                                                                         &      10.00                                                                            &                    73.49                                                                                &    50.00                                         &       93.80              &  46.80  \\ 
\addlinespace[0.1em]
\addlinespace[0.3em]
\ \ + TIR Prompting              &         16.04                                                                         &      14.38                                                                            &                    67.41                                                                                &                    47.66                         &       92.80             &  47.66  \\
\addlinespace[0.1em]
\addlinespace[0.3em]
\ \ + GRPO                       &         20.63                                                                         &      16.25                                                                            &                              70.25                                                                      &        54.38                                  & 87.40                  &   49.78   \\
\addlinespace[0.1em]
\midrule
\addlinespace[0.3em]
\ \ + \ours (ours)                    &          \textbf{22.29}                                                                        &      \textbf{16.88}                                                                            &                  \textbf{72.78}                                                                   &          \textbf{54.53}            & \textbf{89.80}                                 &   \textbf{51.26}    \\ \bottomrule
\end{tabular}
\caption{\label{tab:main_results}
 \textbf{Main experimental results:} 
 we report the passing rate results of \tir Prompting, GRPO, and \ours on five diverse mathematical reasoning benchmarks. We report either the \emph{avg@k} or \emph{pass@k} and $k=\{1,16\}$. 
}
\vspace*{-\baselineskip}
\end{table*}

\subsection{\oursfull}
Adopting turn-level reward assignment, advantage with discounted return, and self-supervised reward shaping, we can obtain the final training objective of \ours by extending from Eq.\ref{eq:grpo} as follows:
\vspace*{-\baselineskip}
\begin{multline}
\mathcal{J}_\text{\ours}(\theta) = \mathbb{E}_{ x , \{y_i\}}
[ \frac{1}{\sum_{i=1}^{G}|y_i|} \sum_{i=1}^{G} \textcolor{red}{\sum_{j=1}^{T_i}} \sum_{t=1}^{|y_{i,j}|} \min  \\
\left( w_{i,j,t} \textcolor{red}{A_{i,j}},  \, \mathrm{clip} ( w_{i,j,t}, 1 - {\varepsilon_{\text{low}}}, 1 + {\varepsilon_{\text{high}}} ) \textcolor{red}{A_{i,j}} \right)] \label{eq:gtpo}
\end{multline}

where $y_{i,j}$ denotes the turn $j$ in sample $i$: $y_{i,j}=t_{i,j}c_{i,j}$, and $T_i$ refers to the total number of turns in the sample $i$.
Note that in Eq.\ref{eq:gtpo},  we still adopt the original formula Eq.\ref{eq:importance_ratio} for the importance ratio $w_{i,j,t}$ and normalize the objective loss by the total number of tokens $\sum_{i=1}^{G}|y_i|$. We also conduct an overhead analysis of \ours in Appendix \ref{sec:appendix_overhead}.

\section{Experiments}
\subsection{Experimental Setup}\label{sec:evaluation_setup}
\paragraph{Cold-Start SFT Training.} 
Following existing work, we first constructed our cold-start dataset based on ReTool-SFT \citep{feng2025retoolreinforcementlearningstrategic}. Specifically, we extracted problems from ReTool-SFT, distilled \tir trajectories from DeepSeek-R1 \citep{deepseekr1} with OpenHands \citep{wang2025openhands} as the scaffold, and conducted rejection sampling by filtering out trajectories whose final answers were incorrect. In the end, our cold-start dataset contains 1.2K problem-trajectory pairs, and we performed SFT on Qwen2.5-7B-Instruct \citep{qwen2025qwen25technicalreport} for three epochs.

\paragraph{RL Training.} 
We used DAPO-17K \citep{yu2025dapo} as our RL training dataset and implemented \ours by extending SkyRL \citep{cao2025skyrl}.
For self-supervised reward shaping in \ours, we used Amazon Titan Text Embeddings V2\footnote{\url{https://docs.aws.amazon.com/bedrock/latest/userguide/titan-embedding-models.html}} to embed the generated code and calculate the similarities between embedding vectors. 
During training, we utilized the AdamW optimizer with an initial learning rate of 1e-6. 
We defined the maximum sequence length for each turn as 8192 tokens, the maximum number of turns as $3$, and the mini-batch size as 1024.
Note that we set the KL coefficient to $0.0$.

\paragraph{Evaluation.} To evaluate the effectiveness of \ours, our evaluation focuses on mathematical reasoning tasks in five diverse math reasoning benchmarks: AIME 2024, AIME 2025\footnote{\url{https://huggingface.co/datasets/AI-MO/aimo-validation-aime}}, MATH 500 \citep{lightman2023lets}, AMC 23\footnote{\url{https://huggingface.co/datasets/zwhe99/amc23}}, and SVAMP \citep{patel2021nlpmodelsreallyable}.
We provide the detailed evaluation settings in Appendix \ref{sec:appendix_temp}.
We considered the following three baselines in our experiments:

\begin{itemize}
    \item \textbf{Simple Prompting:} Chain-of-thought reasoning without any tool calling.
    \item \textbf{\tir Prompting:} Directly using the OpenHands scaffold to generate TIR trajectories without any RL training.
    \item \textbf{GRPO Training for \tir:} Following existing work, to apply GRPO for TIR tasks, we performed cold-start SFT and then used GRPO to further finetune the model. To make a fair comparison, the hyperparameter settings and datasets are the same as fine-tuning \ours.
\end{itemize}

\begin{figure*}
\centering
\begin{minipage}{.66\textwidth}
    \centering
    \includegraphics[width=\linewidth]{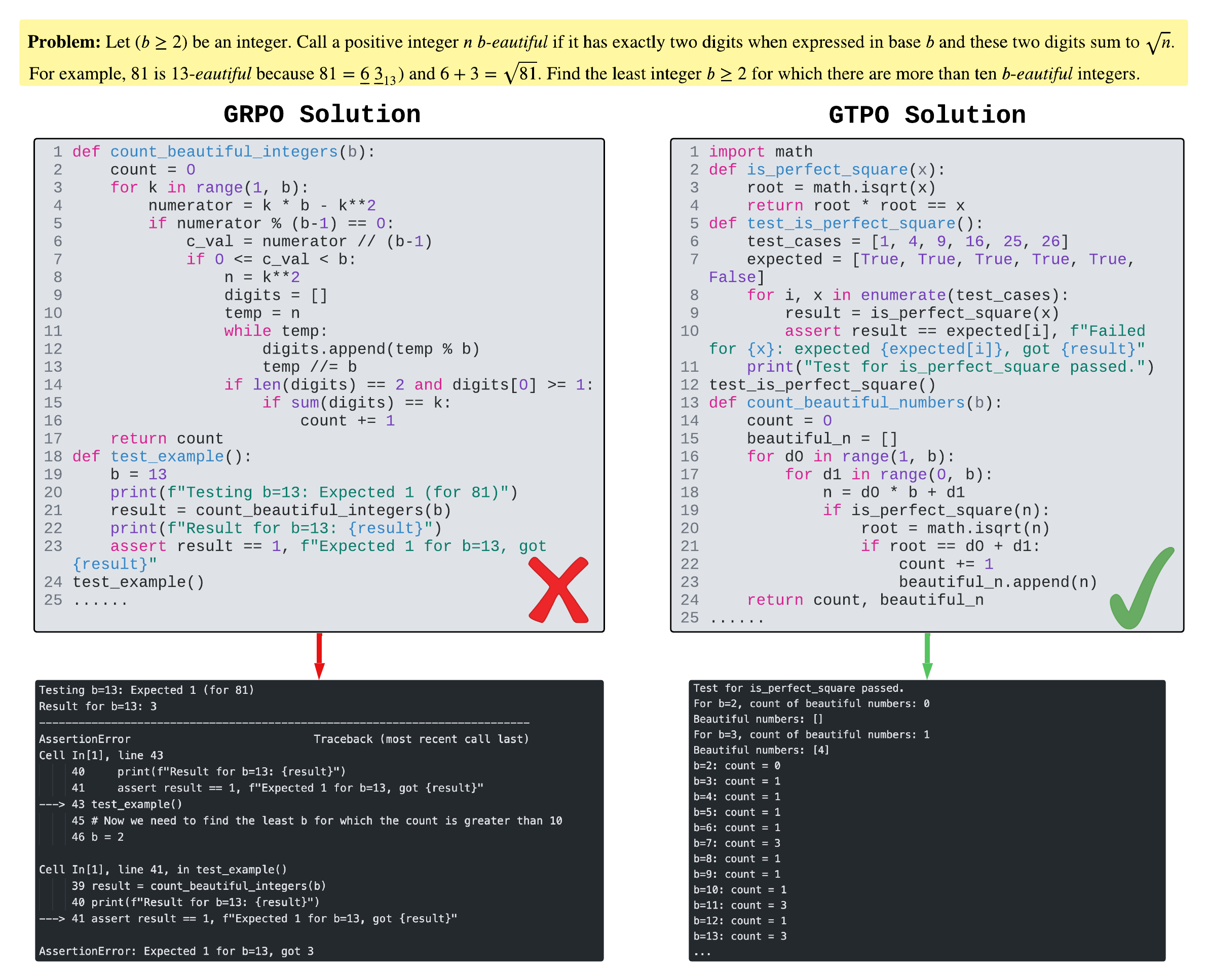}
    \caption{
    \textbf{Qualitative example:}
    We demonstrated an AIME24 example task to compare the distinct coding patterns of GRPO and \ours.
    Qwen2.5-7B-Instruct trained with \ours can write correct code along with accurate tests that thoroughly validate the code correctness, while Qwen2.5-7B-Instruct trained with GRPO fails to solve the problem.
    }
    \label{fig:qualitative_example_aime24}
\end{minipage}%
\hspace{0.01\linewidth}
\begin{minipage}{.32\textwidth}
    \centering
    \includegraphics[width=0.85\linewidth]{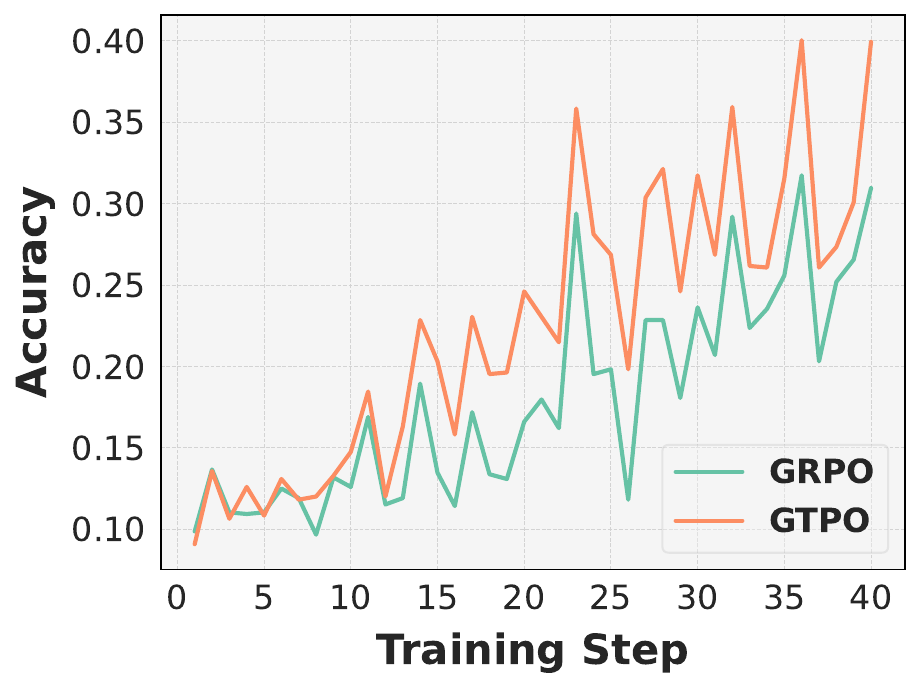}
    \caption{Training accuracy curves of GRPO and \ours under the same experimental setup and training datasets.}
    \label{fig:training_acc}

    \centering
    \includegraphics[width=0.85\linewidth]{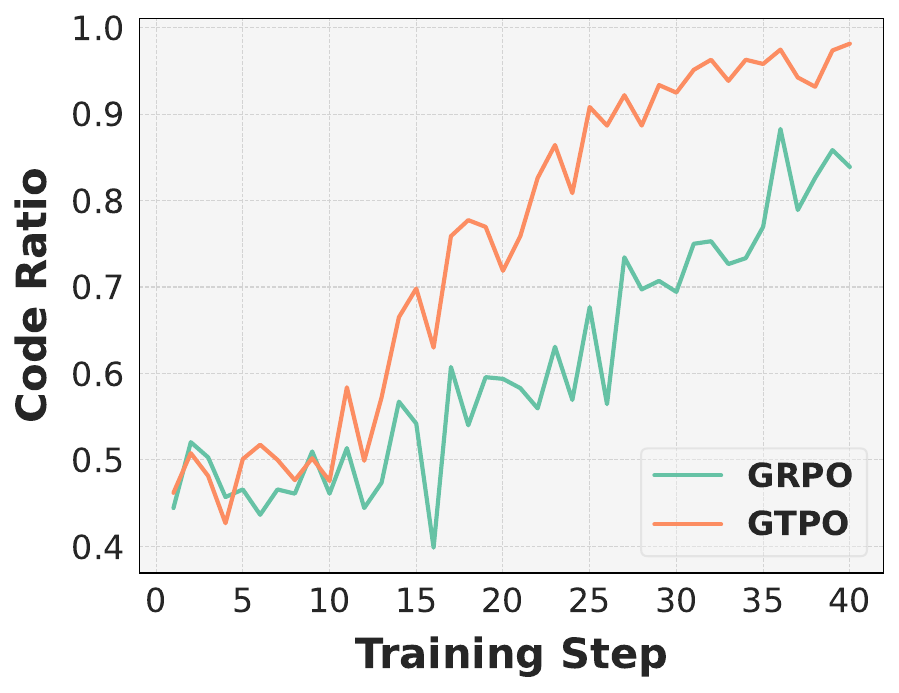}
    \caption{Code ratio curves of GRPO and \ours during training: the code ratio refers to the percentage of rollout trajectories that contain some code content in reasoning.}
    \label{fig:training_code_ratio}
\end{minipage}
\vspace*{-0.8\baselineskip}
\end{figure*}

\subsection{Experimental Results}\label{sec:eval_result}
From Table \ref{tab:main_results}, we observed the following: 
first, directly prompting a conventional language model with TIR is not sufficient, increasing the average passing rate only from 46.80\% (\ie simple prompting) to 47.66\%. 
This observation indicates the need of finetuning language models for TIR to solve complex reasoning tasks. 
Secondly, we found that finetuning with our \ours consistently pushed the performance across all benchmarks, achieving $51.26\%$ passing rate on average. 
This result outperformed GRPO ($49.78\%$) by $3.0\%$ relative improvement, with more significant performance gains in AIME 2024,  MATH 500, and SVAMP. 
Note that we maintained the same training setup between GRPO and GTPO to ensure a fair comparison. 

\begin{table*}[!htb]
    \begin{minipage}{.57\linewidth}
        \centering
        \resizebox{\linewidth}{!}{
        \begin{tabular}{@{}cccccc@{}}
        \toprule
        \begin{tabular}[c]{@{}c@{}}\textbf{Turn-level}\\ \textbf{Reward}\end{tabular}        & \begin{tabular}[c]{@{}c@{}}\textbf{Adv. w/}\\ \textbf{Disct.
        Ret.}\end{tabular}        & \begin{tabular}[c]{@{}c@{}}\textbf{Reward}\\ \textbf{Shaping}\end{tabular}        & \begin{tabular}[c]{@{}c@{}}\textbf{MATH 500}\\ (\emph{pass@1})\end{tabular} & \begin{tabular}[c]{@{}c@{}}\textbf{AIME 2024}\\ (\emph{avg@16})\end{tabular} & \textbf{Average} \\ \midrule
        {\color{Red} \xmark} & {\color{Red} \xmark} & {\color{Red} \xmark} & 67.09 & 20.63 & 43.86                                               \\
        \addlinespace[0.3em]
        {\color{Green} \CheckmarkBold} & {\color{Red} \xmark} & {\color{Red} \xmark} & 72.15 & 20.21 &  46.18                                              \\
        \addlinespace[0.1em]
        \addlinespace[0.3em]
        {\color{Green} \CheckmarkBold} & {\color{Green} \CheckmarkBold} & {\color{Red} \xmark} & 69.94 & 20.63 & 45.29                                               \\
        \hline 
        {\color{Green} \CheckmarkBold} & {\color{Green} \CheckmarkBold} & {\color{Green} \CheckmarkBold} & \textbf{72.47} & \textbf{21.25} &  \textbf{46.86}                                              \\ \bottomrule
        \end{tabular}
        }
        \caption{\label{tab:ablation_main}
        \textbf{Ablation results of \ours}: 
        we reported the results with Qwen2.5-7B-Instruct when removing major components from GTPO: (i) turn-level reward assignment, (ii) advantage with discount return, and (iii) our reward shaping strategy. 
        }
    \end{minipage}%
    \hspace{0.01\linewidth}
    \begin{minipage}{.42\linewidth}
        \centering
        \small
        \resizebox{\linewidth}{!}{
        \begin{tabular}{@{}cccc@{}}
        \toprule
        \begin{tabular}[c]{@{}c@{}}\textbf{Discounting}\\ \textbf{factor ($\gamma$)}\end{tabular}        & \begin{tabular}[c]{@{}c@{}}\textbf{MATH 500}\\ (\emph{pass@1})\end{tabular} & \begin{tabular}[c]{@{}c@{}}\textbf{AIME 2024}\\ (\emph{avg@16})\end{tabular} & \textbf{Average} \\ \midrule
        $0.5$  & 70.57 & 20.21 & 45.39                                               \\
        \addlinespace[0.1em]
        \addlinespace[0.3em]
        $0.7$ & 72.78 & 17.50 & 45.14  \\
        \addlinespace[0.1em]
        \addlinespace[0.3em]
        $0.9$ & \textbf{72.47} & \textbf{21.25} & \textbf{46.86}   \\
        \addlinespace[0.1em]
        \addlinespace[0.3em]
        $1.0$ & 72.15 & 20.21 & 46.18  \\ \bottomrule
        \end{tabular}
        }
        \caption{\label{tab:ablation_gamma}
        \textbf{\ours with different discounting factors ($\gamma$)}: we reported the results on Qwen2.5-7B-Instruct with $\gamma \in [0.5,1]$, where $\gamma = 1$ refers to no discounting in advantage estimate.
        }
    \end{minipage} 
    \vspace*{-\baselineskip}
\end{table*}

\paragraph{Qualitative analysis.} 
In Figure \ref{fig:qualitative_example_aime24}, we showed an example task from AIME 2024 and the critical distinction between code generated by models trained with GRPO and with \ours. 
This task requires finding the least integer $b\geq 2$ for which there are more than ten $b$-eautiful integers.
An integer $n$ is $b$-eautiful if it has exactly two digits in base $b$ and they sum to $\sqrt{n}$. 
The GRPO-trained model demonstrates a fundamental algorithmic flaw: it attempts to validate test cases wrongly in a post-hoc manner after completing the count operation, resulting in an assertion error when $b=13$ yields only $3$ beautiful integers. 
In contrast, the \ours-trained model implements more robust test case validation directly into the search loop (lines 8-10), allowing it to correctly verify the correctness of intermediate results. 
The \ours model successfully identifies that $b=2$ produces $11$ beautiful integers (as shown in the test output), satisfying the problem's requirements.
This example task underscores how finetuning with \ours can correctly incorporate test-driven validation as an integral part of the solution process, indicating superior reasoning enhancement for language models than GRPO.  

\paragraph{Training Curves.} 
We investigated the training progress of GRPO and \ours in Figure \ref{fig:training_acc}.
Compared with GRPO, \ours demonstrates both higher peak performance (reaching approximately 40\% accuracy) as well as greater volatility.
This observation suggests more aggressive optimization and exploration of GTPO for RL training. 
GRPO exhibits a more conservative learning pattern as we found its accuracy plateauing around 25\% - 30\% in the later stages. 
This persistent performance gap between GRPO and GTPO demonstrates the effective use of reward feedback from our method to improve the model reasoning capabilities.

Figure \ref{fig:training_code_ratio} compares the code ratio curves of GRPO and \ours. 
\ours demonstrates a more aggressive shift toward code-based reasoning, reaching nearly 98\% code ratio by training step 40 with a relatively consistent upward trend. 
In contrast, GRPO exhibits a more conservative learning pattern, plateauing around 85\%, suggesting potential instability or exploration-exploitation trade-offs in its optimization process. 
Compared with GRPO, training with \ours significantly pushes the model to use code as an external tool for \tir tasks.

\paragraph{Ablation Studies.}
We performed ablation studies to investigate the design choices of \ours. 
All the ablation experiments start from Qwen2.5-7B-Instruct after cold-start SFT, following the same settings described in \S\ref{sec:evaluation_setup}. 
First, we conducted an experiment to study the effect of return discounting and self-supervised reward shaping in \ours. 
As shown in Table \ref{tab:ablation_main}, the model trained with both return discounting and self-supervised reward shaping achieves the best evaluation results on MATH 500 and AIME 2024, achieving $46.86\%$ passing rate on average. 
When we removed these components gradually, we observed negative performance impacts with more significant performance drops when removing both turn-level reward and reward shaping strategies. 
These observations highlight the importance of different components in \ours for multi-turn TIR tasks.

\paragraph{Comparison with Additional Baseline.}
To compare \ours with more baselines, we experiment comparing \ours against DAPO and SimpleTIR on Qwen2.5-7B-Instruct, using the same training setup and computational budget in the ablation experiments (\S\ref{sec:appendix_temp}). Following standard practice, we evaluate using MATH 500 (pass@1) and AIME 2024 (avg@16). As shown in Table \ref{tab:discussion_dapo}, while both DAPO and SimpleTIR provide small improvements over GRPO, \ours still achieves the best performance, supporting our claim that \ours offers benefits beyond standard RL objectives and recent tool-RL alternatives.

\begin{table}[htbp]
\centering
\resizebox{0.48\textwidth}{!}{
\begin{tabular}{@{}lccc@{}}
\toprule
\textbf{Qwen2.5-7B-Instruct}              & \multicolumn{1}{c}{\begin{tabular}[c]{@{}c@{}}\textbf{MATH 500}\\ (\emph{pass@1})\end{tabular}} & \multicolumn{1}{c}{\begin{tabular}[c]{@{}c@{}}\textbf{AIME 2024}\\ (\emph{avg@16})\end{tabular}} & \textbf{Average} \\ \midrule
\ \ + GRPO                       &        67.09                                                                          &            20.63                                                                      &                              43.86         \\
\addlinespace[0.1em]
\ \ + DAPO                       &         67.72                                                                         &      20.77                                                                            &                              44.25         \\
\addlinespace[0.1em]
\ \ + SimpleTIR                       &         68.67                                                                         &      20.83                                                                            &                              44.75         \\
\addlinespace[0.1em]
\midrule
\addlinespace[0.3em]
\ \ + \ours (ours)                    &          \textbf{72.47}                                                                        &      \textbf{21.25}            &   \textbf{46.86}    \\ \bottomrule
\end{tabular}
}
\caption{\label{tab:discussion_dapo}
 \textbf{Comparison with additional baselines:} 
 we add DAPO and SimpleTIR in our evaluation. Results show that our turn-level reward formulation provides complementary advantages over existing RL methods and recent tool-RL alternatives.
}
\vspace{-0.2in}
\end{table}

\paragraph{Impact of Discounting Factor.}
We studied the effect of the discounting factor $\gamma$.
In our experiment, we study four different values of the discounting factor: $\gamma=\{0.5, 0.7, 0.9, 1.0\}$. 
As shown in Table \ref{tab:ablation_gamma}, $\gamma = 0.9$ achieves the optimal balance between distant rewards and immediate rewards, outperforming the conventional approach in GRPO (where $\gamma=1.0$). 
When $\gamma < 0.9$, we observed negative performance impacts with a reduction of passing rate up to $1.72\%$. 
These observations demonstrate the importance of the discounting factor for any turn-based RL approaches like GTPO. 

For more experimental results, please refer to Appendix \ref{sec:appendix_embedding}, \ref{sec:appendix_case}, and \ref{sec:appendix_curve}.

\section{Discussion}

\subsection{Generalization across Model Families}
To assess the generality of \ours across architectures and scales, we additionally trained Llama-3.2-3B-Instruct with both GRPO and \ours and evaluated on AIME 2024 (avg@16) and MATH 500 (pass@1). As shown in Table \ref{tab:discussion_generalization_model}, \ours consistently outperforms GRPO, improving accuracy from 21.18\% to 23.01\% on average, supporting our claim that \ours generalizes across diverse LLM families and model sizes.

\begin{table}[htbp]
\centering
\resizebox{0.48\textwidth}{!}{
\begin{tabular}{@{}lccc@{}}
\toprule
\textbf{Llama3.2-3B-Instruct}        & \multicolumn{1}{c}{\begin{tabular}[c]{@{}c@{}}\textbf{MATH 500}\\(\emph{pass@1})\end{tabular}}      & \multicolumn{1}{c}{\begin{tabular}[c]{@{}c@{}}\textbf{AIME 2024}\\ (\emph{avg@16})\end{tabular}}   & \textbf{Average} \\ \midrule
\ \ + GRPO             &      37.97          &         4.38                                                                                                                                                     &                              21.18         \\
\addlinespace[0.1em]
\midrule
\addlinespace[0.3em]
\ \ + \ours (ours)          &      \textbf{39.56}          &          \textbf{6.46}                                                                                    &   \textbf{23.01}    \\ \bottomrule
\end{tabular}
}
\caption{\label{tab:discussion_generalization_model}
 \textbf{Generalization across model families:} 
 we validated the generalizability of \ours across model families and sizes. On Llama-3.2-3B-Instruct, \ours consistently outperforms GRPO by 8.6\% on average.
}
\vspace*{-\baselineskip}
\end{table}

\subsection{Generalization to Non-Math Domains}
We extended our evaluation beyond mathematical reasoning to assess \ours on two additional task families: commonsense reasoning and program synthesis. For commonsense reasoning, we evaluated on GPQA Diamond~\citep{rein2023gpqagraduatelevelgoogleproofqa} and BBEH-mini~\citep{kazemi2025bigbenchextrahard}, using tool-integrated reasoning for each problem. For program synthesis, we use EvalPlus~\citep{evalplus}, including HumanEval and MBPP, which is a natural fit since TIR-trained models tend to develop stronger Python-based problem-solving skills.

As shown in Table \ref{tab:discussion_generalization_task}, across all four benchmarks, \ours consistently outperforms GRPO with 3.9\% relative improvement on average. This shows that \ours can generalize beyond mathematical reasoning and provide gains for non-math tasks.

\begin{table}[htbp]
\centering
\resizebox{0.48\textwidth}{!}{
\begin{tabular}{@{}lccccc@{}}
\toprule
\textbf{Qwen2.5-7B}              & \multicolumn{1}{c}{\begin{tabular}[c]{@{}c@{}}\textbf{GPQA}\\ (\emph{pass@1})\end{tabular}} & \multicolumn{1}{c}{\begin{tabular}[c]{@{}c@{}}\textbf{BBEH}\\ (\emph{pass@1})\end{tabular}} & \multicolumn{1}{c}{\begin{tabular}[c]{@{}c@{}}\textbf{HumanEval}\\ (\emph{pass@1})\end{tabular}} & \multicolumn{1}{c}{\begin{tabular}[c]{@{}c@{}}\textbf{MBPP}\\ (\emph{pass@1})\end{tabular}} & \textbf{Avg.} \\ \midrule
\ \ + GRPO                       &         21.7                                                                         &      4.1   &         77.4                                                                         &      81.0                                                                            &                              46.1         \\
\addlinespace[0.1em]
\midrule
\addlinespace[0.3em]
\ \ + \ours (ours)                   &          \textbf{24.2}                                                                        &      \textbf{4.4}            &   \textbf{81.7}   &          \textbf{81.2}                                                                        &      \textbf{47.9} \\ \bottomrule
\end{tabular}
}
\caption{\label{tab:discussion_generalization_task}
 \textbf{Generalization to non-math domains:} 
 we evaluate the generalizability of \ours to commonsense reasoning and program synthesis tasks. \ours maintains stable improvements over GRPO with 3.9\% relative improvement on average. 
}
\vspace*{-\baselineskip}
\end{table}

\subsection{Robustness of Embedding Choices}
We study the robustness of the selection of different embedding models for self-supervised reward shaping in \ours. Specifically, we computed the code similarity between correct and incorrect trajectories using three embedding models: Amazon Titan Text Embeddings V2 (used in our main experiments), OpenAI text-embedding-3-large\footnote{\url{https://developers.openai.com/api/docs/models/text-embedding-3-large}}, and Nomic Embed Text V1.5\footnote{\url{https://huggingface.co/nomic-ai/nomic-embed-text-v1.5}}, which are trained by different companies with diverse dimensions ranging from 768 to 3072. We then calculated the cross-model correlation of these code similarity scores.

As shown in Table \ref{tab:discussion_embedding}, we observed very strong Pearson (>0.75) and Spearman (>0.84) correlations across all models. Such high correlations indicate that code similarity scores are preserved across embedding families, suggesting that the reward shaping of \ours is robust to embedding choice, as our shaping mechanism depends primarily on the similarity structure between code in trajectories.

\begin{table}[htbp]
\centering
\resizebox{0.48\textwidth}{!}{
\begin{tabular}{@{}lccc@{}}
\toprule
\multicolumn{4}{c}{\textbf{Pearson Correlation}} \\ 
\midrule
 & \textbf{Amazon Titan} & \textbf{Nomic} & \textbf{OpenAI} \\ 
\midrule
\textbf{Amazon Titan}            & 1.0000 & 0.7834 & 0.7526 \\
\textbf{Nomic}        & 0.7834 & 1.0000 & 0.8450 \\
\textbf{OpenAI}   & 0.7526 & 0.8450 & 1.0000 \\
\midrule
\addlinespace[0.3em]
\multicolumn{4}{c}{\textbf{Spearman Correlation}} \\ 
\midrule
 & \textbf{Amazon Titan} & \textbf{Nomic} & \textbf{OpenAI} \\ 
\midrule
\textbf{Amazon Titan}            & 1.0000 & 0.8429 & 0.8500 \\
\textbf{Nomic}        & 0.8429 & 1.0000 & 0.9071 \\
\textbf{OpenAI}   & 0.8500 & 0.9071 & 1.0000 \\
\bottomrule
\end{tabular}
}
\caption{\label{tab:discussion_embedding}
\textbf{Correlation comparison across embedding models:} 
we report both Pearson and Spearman correlations between Amazon Titan Text Embeddings V2, Nomic Embed Text V1.5, and OpenAI text-embedding-3-large, showing strong agreement and robustness across different embedding models.
}
\vspace{-0.2in}
\end{table}

\subsection{Improvement over Tool Correctness}
While Figure \ref{fig:training_code_ratio} shows that \ours can increase tool usage frequency (\ie code ratio) effectively, we conduct an additional experiment to show that \ours can also generate more “useful tool calls”. Because evaluating step-by-step logic errors automatically is intractable without intermediate oracles, we evaluated runtime error rates as a proxy for tool correctness. Specifically, runtime success reflects whether tool invocations are syntactically valid and executable, which is an important prerequisite for effective TIR. To this end, we compare the percentage of tool calls that execute without runtime errors across benchmarks from Qwen2.5-7B-Instruct trained with GRPO and \ours. As shown in Table \ref{tab:discussion_tool_correctness}, \ours improves tool correctness by nearly 3.0\% on average, demonstrating that our turn-level reward successfully teaches the model to write more reliable, executable code rather than simply spamming tool calls.

\begin{table}[htbp]
\centering
\resizebox{0.48\textwidth}{!}{
\begin{tabular}{@{}lcccccc@{}}
\toprule
\textbf{Method} & \textbf{AIME24} & \textbf{AIME25} & \textbf{MATH500} & \textbf{AMC23} & \textbf{SVAMP} & \textbf{Avg} \\
\midrule
GRPO  & 69.23 & 67.57 & 75.80 & \textbf{72.17} & 99.05 & 76.77 \\
\addlinespace[0.1em]
\midrule
\addlinespace[0.3em]
\ours  & \textbf{74.45} & \textbf{73.19} & \textbf{82.07} & 69.49 & \textbf{99.54} & \textbf{79.75} \\
\bottomrule
\end{tabular}
}
\caption{\label{tab:discussion_tool_correctness}
\textbf{Improvement over tool correctness:}
compared to GRPO, \ours consistently improves tool correctness across benchmarks by nearly 3.0\% on average. This result suggests that GTPO’s turn-level reward formulation encourages the model to produce more reliable and executable tool calls.
}
\vspace{-0.2in}
\end{table}

\section{Conclusion}
In this work, we addressed the challenge of training language models for multi-turn Tool-Integrated Reasoning through RL. 
Our solution, \textbf{\oursfull (\ours)}, introduces turn-level reward functions with rule-based rewards for individual turns and turn-level reward discounting for advantage calculation, overcoming trajectory-level reward limitations. 
Additionally, our reward shaping technique uses self-supervision signals from generated code to densify sparse binary rewards, improving learning efficiency. 
Empirical results demonstrate that \ours achieves 3.0\% relative improvement on average over GRPO on five diverse math reasoning benchmarks. \ours also improves \grpo by 3.9\% on commonsense reasoning and program synthesis tasks, demonstrating its generalizability to non-math domains. Furthermore, \ours incurs negligible overhead, setting it as a new advanced RL technique to improve model reasoning in the real world.

\section*{Limitations}

While \ours has proven effective through extensive evaluation in the paper, our experiments are restricted to models no larger than 7B parameters due to the computation budget. It is prohibitively expensive to perform
large-scale RL experiments for LLMs, and unfortunately, we do not have enough resources to demonstrate the impact of \ours on larger models. In addition, this work has mainly focused on Tool-Integrated Reasoning tasks, but the idea of \ours can be broadly applicable for improving models’ reasoning capability in general multi-turn scenarios such as real-world software engineering tasks, which we leave to future work.

\bibliography{custom}

\appendix

\begin{figure*}[!h]
\centering
\includegraphics[width=\linewidth]{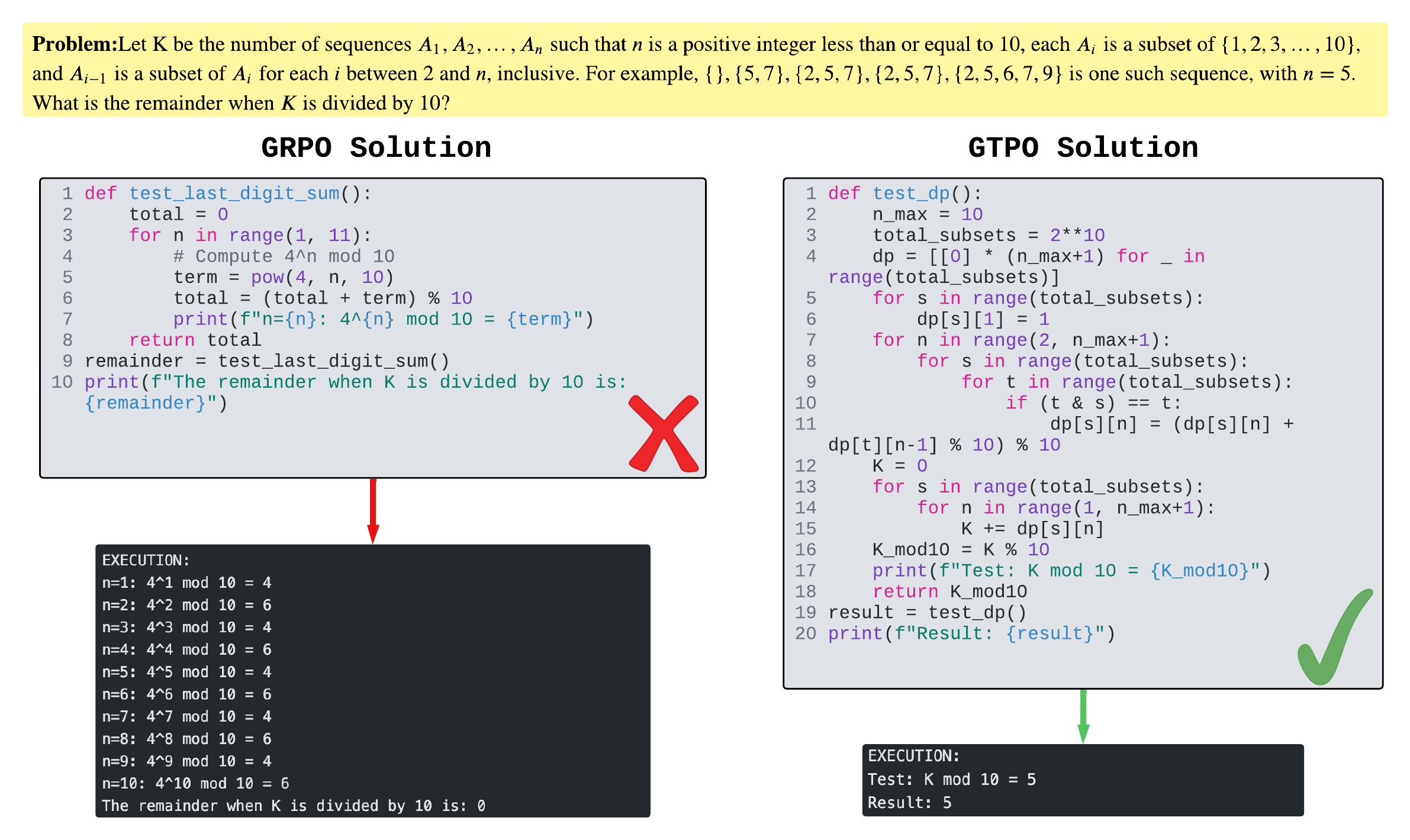}
\caption{
An AMC23 example to compare the distinction in generation samples between GRPO and \ours. 
}
\label{fig:qualitative_example_amc23}
\vspace*{-\baselineskip}
\end{figure*}

\newpage
\section{Appendix}
\label{sec:appendix}

\subsection{Comparison of \ours and related works}\label{sec:appendix_comparison}
Different from prior works, we introduced \ours by \textbf{reformulating multi-turn TIR as a turn-level MDP with a specific, tool-aware reward structure} and integrating this with group-based policy optimization. This reformulation is simple yet critical to address a fundamental modeling issue in prior TIR works: these methods continue to treat the entire tool-using trajectory as a single action with a sparse outcome reward. Below, we discussed in detail the differences between \ours and prior work along the following dimensions: turn-level reward assignment, discounted return for advantage calculation, and self-supervised reward shaping for negative trajectories.

\subsubsection{Turn-level reward assignment}\label{sec:appendix_comparison_1}
GRPO-style methods in TIR treat the entire multi-turn trace as one action with a single trajectory-level reward (e.g., final correctness), broadcasting the same scalar advantage to all tokens. Our approach (\ours) explicitly redefines the decision process at the turn level: each turn is an action whose reward is computed from tool feedback and signals (e.g., format correctness, presence, and quality of tool calls). This is not just a cosmetic tweak of GRPO’s reward; \textbf{it changes the underlying state–action structure from a bandit over full trajectories to a sequential decision process over turns, which is tailored to multi-turn TIR}. Our ablations (Table \ref{tab:ablation_main}) show that this reformulation alone yields clear gains over trajectory-level GRPO under the same data and infrastructure.

\subsubsection{Discounted return for advantage calculation}\label{sec:appendix_comparison_2}
GRPO/GSPO/DAPO typically use group-normalized, trajectory-level rewards that ignore temporal structure: every token in a trajectory receives the same normalized advantage scalar. Our approach (\ours) instead computes \textbf{discounted reward-to-go at the turn level}, so terminal success or failure propagates backward in a degraded way. To our knowledge, prior GRPO-style methods for TIR do not (i) define a turn-level MDP and simultaneously (ii) use turn-wise discounted returns within the group objective. Our ablations (Tables \ref{tab:ablation_main}–\ref{tab:ablation_gamma}) show that removing discounted turn-level return degrades performance and that the discount factor $\gamma$ has a clear “good” range.

\subsubsection{Self-supervised reward shaping for negative trajectories}\label{sec:appendix_comparison_3}
Prior TIR RL work (e.g., ReTool, ToRL, Search-R1) typically uses binary or near-binary outcome rewards, treating all incorrect trajectories as equally bad and providing little signal when all samples in a group fail. \ours introduces self-supervised shaping for failed trajectories: we assign partial rewards based on the similarity between their code and code from successful trajectories. \textbf{This recovers informative signal from “almost correct” code and exploits the fact that code is a concise, verifiable proxy for intermediate reasoning quality.} Table \ref{tab:ablation_embedding} shows that this shaping outperforms both purely binary rewards and simpler string-matching baselines.

The combined three designs target the core RL challenge: sparse, delayed rewards and poor temporal credit assignment~\citep{pignatelli2024surveytemporalcreditassignment}. \ours simultaneously (1) changes the MDP granularity to turns, (2) propagates credit via discounted returns over turns, and (3) densifies supervision inside failed trajectories using code-based self-supervision. To our knowledge, \ours is the first method to instantiate this specific turn-level, discounted, self-supervised RL formulation for multi-turn TIR, and our experiments and ablations consistently show empirical gains over standard GRPO-style baselines.

\subsection{Turn-level Format Reward Design}\label{sec:appendix_reward}
In practice, considering the nature of \tir tasks, we focus on two major format requirements: (1) the format of tool calling must be correct, and (2) there must exist at least one tool call throughout the trajectory. Specifically, we assign $r_{\text{format}_{i,j}} = -0.1$ when $y_j$ contains any invalid tool calls. To further ensure that at least one tool call happens throughout the trajectory, we directly demand the first turn $y_1$ to contain tool calls: we assign $r_{\text{format}_{i,1}} = -0.1$ if $y_1$ does not include any tool calls. The reason is that, based on our observation, for all the trajectories that contain tool calls, models will always make tool calls in the first turn. Following DAPO~\cite{yu2025dapo}, we use the answer format "Answer:" throughout all the evaluations.

\begin{algorithm*}[t]
\caption{Group Turn Policy Optimization (GTPO)}
\label{alg:gtpo}
\begin{algorithmic}[1]
\Require Policy $\pi_{\theta}$, reference policy $\pi_{\theta_{\text{old}}}$, dataset $\mathcal{D}$, group size $G$, discount $\gamma$, clip $\epsilon_{\text{low}}, \epsilon_{\text{high}}$, shaping cap $\alpha$, learning rate $\eta$
\For{each update step}
  \State Sample prompt $x \sim \mathcal{D}$
  \State Roll out $G$ trajectories $\{y_i\}_{i=1}^G \sim \pi_{\theta_{\text{old}}}(\cdot \mid x)$, where $y_i = \{y_{i,1},\dots,y_{i,T_i}\}$ and $y_{i,j}=(t_{i,j},c_{i,j})$
  \State Let $\mathcal{P} \gets \{ i \mid \text{final answer of } y_i \text{ is correct}\}$
  \For{$i=1,\dots,G$}
    \For{$j=1,\dots,T_i$}
      \State $\displaystyle r_{{\text{format}}_{i,j}} \gets 
      \begin{cases}
      -0.1 & \text{if } y_{i,j} \text{ has any format error}\\
      0 & \text{otherwise}
      \end{cases}$
      \State $\displaystyle r_{{\text{acc}}_{i,j}} \gets
      \begin{cases}
      0 & j < T_i\\
      1 & (j=T_i) \wedge (i\in\mathcal{P})\\
      \frac{\alpha}{|\mathcal{P}|}\sum\limits_{p\in\mathcal{P}}
      \mathrm{sim}\!\Big(\!\bigoplus_{m<j} c_{i,m}, \bigoplus_{m<j} c_{p,m}\!\Big)
      & (j=T_i) \wedge (i\notin\mathcal{P})
      \end{cases}$
      \State $r_{i,j} \gets r_{{\text{acc}}_{i,j}} + r_{{\text{format}}_{i,j}}$
      \State $\displaystyle R_{i,j} \gets \sum_{m=j}^{T_i} \gamma^{\,m-j} \, r_{i,m}$
    \EndFor
  \EndFor
  \State Compute global mean $\mu$ and std $\sigma$ over all $\{R_{i,j}\}$ (all $i,j$)
  \State $A_{i,j} \gets (R_{i,j}-\mu)/(\sigma+\delta)$
  \State $\displaystyle \mathcal{L}(\theta) \gets - \frac{1}{\sum_i |y_i|}\sum_{i=1}^G\sum_{j=1}^{T_i}\sum_{t\in y_{i,j}}
  \min\!\Big(w_{i,j,t} A_{i,j},\, \mathrm{clip}(w_{i,j,t},1-\epsilon_{\text{low}},1+\epsilon_{\text{high}})A_{i,j}\Big)$
  \State where $\displaystyle w_{i,j,t}=\frac{\pi_\theta(y_{i,j,t}\mid x,\text{history})}{\pi_{\theta_{\text{old}}}(y_{i,j,t}\mid x,\text{history})}$
  \State Update $\theta \leftarrow \theta - \eta \nabla_\theta \mathcal{L}(\theta)$
\EndFor
\end{algorithmic}
\end{algorithm*}

\subsection{Overhead Analysis}\label{sec:appendix_overhead}
\ours introduces minimal and negligible computational overhead compared to GRPO. Firstly, the overhead of turn-level reward calculation is minimal. Both the format and outcome rewards rely solely on lightweight string-matching operations; thus, extending them from the trajectory level to the turn level does not introduce any meaningful additional cost. Likewise, the overhead of our self-supervised reward shaping is negligible: it only requires computing a single embedding per sampled trajectory, which takes on the order of seconds and is insignificant compared to the cost of generating full trajectories.

\subsection{Additional Evaluation Settings}\label{sec:appendix_temp}
For benchmarks where we report \textit{avg@16} performance, including AIME 2024, AIME 2025, and AMC 2023, we set the sampling temperature to be 0.6. For other benchmarks where we report \textit{pass@1} performance, including MATH500 and SVAMP, we set a lower sampling temperature of 0.2 for more stable evaluation results. In the evaluation, we define the maximum sequence length for each turn as 8192 tokens and the maximum number of turns as 10, allowing models for more exploration. Because the RL training set we use (\ie DAPO-17K~\cite{yu2025dapo}) only includes problems whose answer is a single integer, we filter out all the problems from these benchmarks whose ground truth is not a single integer. In the main experiments in Table \ref{tab:main_results}, both GRPO and \ours checkpoints we evaluate have been trained for 40 steps. Limited by computational resources, for all the other experiments, we evaluate checkpoints that have been trained for 30 steps.

\subsection{Impact of Reward Shaping Strategy}\label{sec:appendix_embedding}
We evaluated different design choices for \ours reward shaping. 
Specifically, we compared character-based sequence matching via \texttt{Difflib} \footnote{\url{https://docs.python.org/3/library/difflib.html}} \citep{wei2025swerladvancingllmreasoning} and embedding-based similarity \citep{zhang2024code}.
We also varied the content scope used to compute similarity, contrasting entire trajectories (natural language + code) with code-only inputs.
As shown in Table \ref{tab:ablation_embedding}, representing trajectory correctness using only code components yields the largest gains. In particular, embedding-based similarity computed on code-only content achieves the best performance. 
This finding suggests that, for TIR tasks, code provides a concise and reliable feedback signal for steering model reasoning, replacing the conventional natural language data.

\begin{table}[t]
\centering
\resizebox{\columnwidth}{!}{
\begin{tabular}{@{}ccccc@{}}
\toprule
\begin{tabular}[c]{@{}c@{}}\textbf{Scoring}\\ \textbf{method}\end{tabular}        & \begin{tabular}[c]{@{}c@{}}\textbf{Sample}\\ \textbf{content}\end{tabular}        & \begin{tabular}[c]{@{}c@{}}\textbf{MATH 500}\\ (\emph{pass@1})\end{tabular} & \begin{tabular}[c]{@{}c@{}}\textbf{AIME 2024}\\ (\emph{avg@16})\end{tabular} & \textbf{Avg.} \\ \midrule
\addlinespace[0.3em]
\texttt{Difflib} & Code & 71.52    &  21.25  & 46.39                                                    \\
\addlinespace[0.1em]
\addlinespace[0.3em]
\texttt{Difflib} & Trajectory & 72.15  & 18.75 & 45.45           \\
\hline
\addlinespace[0.1em]
Embedding & Code  & \textbf{72.47}  &   \textbf{21.25}  & \textbf{46.86}                                             \\
\bottomrule
\end{tabular}
}
\caption{\label{tab:ablation_embedding}
\textbf{\ours by reward shaping strategies}: 
we change the reward shaping strategy by the scoring method (using embedding model or \texttt{Difflib}) and the data sample content (code only or the whole trajectory). 
}
\vspace*{-\baselineskip}
\end{table}

\subsection{Additional Case Studies}\label{sec:appendix_case}
In Figure \ref{fig:qualitative_example_amc23}, we show an example from AMC23, illustrating the distinct problem-solving approaches between GRPO and \ours models in tackling combinatorial counting problems.
The task requires counting sequences of subsets with specific containment properties modulo $10$ - a problem that demands careful handling of the exponential growth in possibilities. 
The GRPO solution attempts a direct computational approach using dynamic programming with memoization, but critically fails to properly manage the modular arithmetic.
Specifically, it computes the full count first and only applies the modulo operation at the end, leading to integer overflow issues that produce an incorrect result of 0. 
In contrast, the \ours solution demonstrates superior algorithmic insight by maintaining the modulo 10 constraint throughout the computation within its dynamic programming table, preventing overflow and correctly identifying the answer as 5.

\subsection{Training Curves}\label{sec:appendix_curve}

\begin{figure}[h!]
\centering
\includegraphics[width=0.9\linewidth]{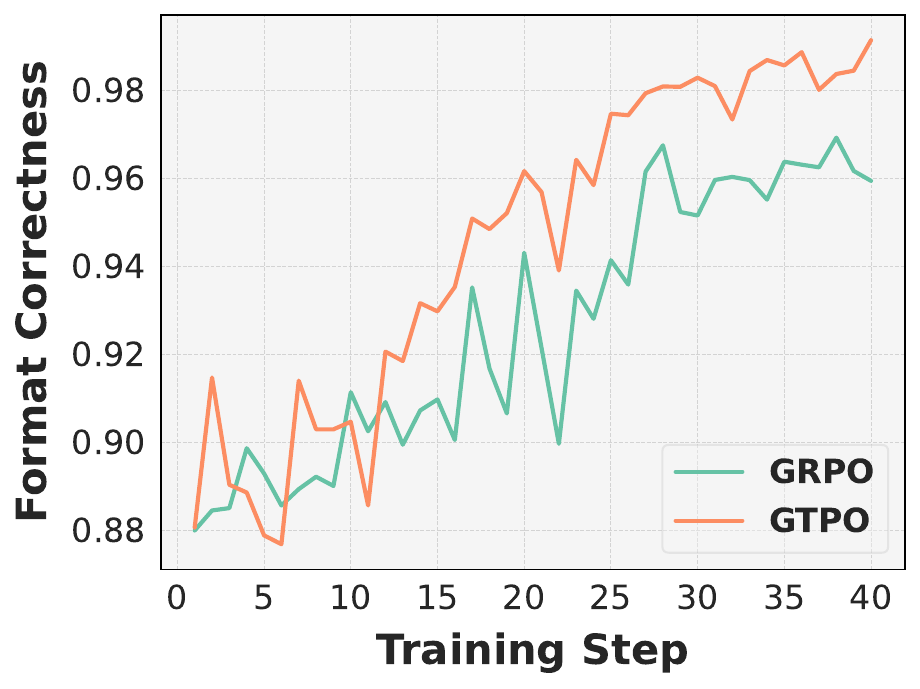}
\caption{Format correctness curves of GRPO and \ours: 
the format correctness metric refers to the percentage of rollout trajectories that do not include any format-based errors (e.g. generated code blocks).}
\label{fig:training_format_correctness}
\vspace*{-\baselineskip}
\end{figure}

Figure \ref{fig:training_format_correctness} demonstrates the format correctness curves of GRPO and \ours. 
\ours exhibits superior performance throughout the training process, achieving a robust improvement to around 99\% by training step 40.
In contrast, GRPO shows more volatile behavior, particularly evident in the dramatic spike and subsequent drop around training steps 20-25. 
While GRPO eventually recovers and stabilizes around 97\% by the end of training, it consistently underperforms \ours by approximately 2-3 percentage points in the later stages. 

\subsection{Additional Results}

\paragraph{Scaling with different trajectory turns}.
We further conducted ablation study by changing the maximum number of turns in generated trajectories. As shown in Table \ref{tab:ablation_maxiter}, the performance of \ours improves steadily when increasing the number of maximum turns. This observation shows that \ours scales well with the trajectory length in multi-turn \tir reasoning.

\begin{table}[!h]
\centering
\small
\begin{tabular}{@{}cccc@{}}
\toprule
\begin{tabular}[c]{@{}c@{}}\textbf{Max}\\ \textbf{turns $\mathcal{T}$}\end{tabular}        & \begin{tabular}[c]{@{}c@{}}\textbf{MATH 500}\\ (\emph{pass@1})\end{tabular} & \begin{tabular}[c]{@{}c@{}}\textbf{AIME 2024}\\ (\emph{avg@16})\end{tabular} & \textbf{Average} \\ \midrule
$1$  & 71.20 & 19.38 & 45.29                                               \\
\addlinespace[0.1em]
\addlinespace[0.3em]
$2$  & 69.62 & 20.21 & 44.92                                               \\
\rowcolor{blue!40}
\addlinespace[0.1em]
\addlinespace[0.3em]
$3$  & \textbf{72.47} & \textbf{21.25} & \textbf{46.86}                                                \\ \bottomrule
\end{tabular}
\caption{\label{tab:ablation_maxiter}
\textbf{\ours by different maximum trajectory turns:} 
We conducted experiments with \ours where we changed the maximum number of turns in rollout trajectories $\mathcal{T}=\{1,2,3\}$ during training. 
}
\end{table}

\end{document}